\documentclass[10pt,journal,compsoc]{IEEEtran}
\usepackage{xcolor,soul,framed} %,caption

\colorlet{shadecolor}{yellow}
\usepackage[pdftex]{graphicx}
\graphicspath{{../pdf/}{../jpeg/}}
\DeclareGraphicsExtensions{.pdf,.jpeg,.png}

\usepackage[cmex10]{amsmath}
\usepackage{mdwmath}
\usepackage{mdwtab}
\usepackage{eqparbox}
\newcolumntype{P}[1]{>{\centering\arraybackslash}p{#1}}
\usepackage{url}
\usepackage[caption=false]{subfig}
\usepackage{array, booktabs}
\usepackage{amsmath,amssymb,amsfonts}
\usepackage{mathtools}
\usepackage{multirow}
\usepackage{makecell}
\usepackage{algorithmic}
\usepackage[]{algorithm2e}
\usepackage{pifont}
\usepackage{verbatim}
\usepackage{subfig}
\usepackage{tabularx}
\usepackage{textcomp}
\usepackage{float}
\usepackage{cite}
\usepackage[normalem]{ulem} %to strike the words
\begin{document}
%\bstctlcite{IEEEexample:BSTcontrol}
%\markboth{IEEE TRANSACTIONS ON CIRCUITS AND SYSTEM FOR VIDEO TECHNOLOGY, VOL.~60, NO.~12, APRIL~2020
%}

%jchoi: title - Attract -> Attracting ...
%             - A Method to Attract ...
\title{LTC-GIF: Attracting More Clicks on Feature-length Sports Videos}

% \title{Thumbnail GIF: Attract More Clicks on Feature-length Sports Videos}

% Lightweight Method to Generate Thumbnails and Animated GIF from Feature-length Videos

% Lightweight Client-driven GIF Generation Framework for Feature-length Videos

\author{
% \thanks{This work was supported by a National Research Foundation of Korea (NRF) grant funded by the Korean Government grant No. 2019R1A2C1010476. This work was also supported by Ministry of Science and ICT (MSIT) Korea, under the Information Technology Research Center (ITRC) support program IITP-2020-2017-0-01630 supervised by the Institute for Information \& communications Technology Promotion (IITP).}

Ghulam~Mujtaba, Jaehyuk Choi,~\IEEEmembership{Member,~IEEE}, and Eun-Seok Ryu,~\IEEEmembership{Senior Member,~IEEE}
\IEEEcompsocitemizethanks{ \IEEEcompsocthanksitem G. Mujtaba is with the Department of Computer Engineering, Gachon University, Seongnam, Republic of Korea.\protect\\
E-mail: mujtaba@gachon.ac.kr.
\IEEEcompsocthanksitem J. Choi is with the Department of Software, Gachon University, Seongnam, Republic of Kore.\protect\\
E-mail: jchoi@gachon.ac.kr.
\IEEEcompsocthanksitem E.-S. Ryu is with the Department of Computer Education, Sungkyunkwan University (SKKU), Republic of Korea.\protect\\
E-mail: esryu@skku.edu.
}

}

\markboth{}
{Mujtaba, Choi and Ryu: LTC-GIF: Attracting More Clicks on Feature-length Sports Videos}

% ====================================================================
\maketitle

% === ABSTRACT ====================================================================
% =================================================================================
\begin{abstract}
This paper proposes a lightweight method to attract users and increase views of the video by presenting personalized artistic media -- i.e, static thumbnails and animated GIFs. This method analyzes lightweight thumbnail containers (LTC) using computational resources of the client device to recognize personalized events from full-length sports videos. In addition, instead of processing the entire video, small video segments are processed to generate artistic media. This makes the proposed approach more computationally efficient compared to the baseline approaches that create artistic media using the entire video. The proposed method retrieves and uses thumbnail containers and video segments, which reduces the required transmission bandwidth as well as the amount of locally stored data used during artistic media generation. When extensive experiments were conducted on the Nvidia Jetson TX2, the computational complexity of the proposed method was 3.57 times lower than that of the SoA method. In the qualitative assessment, GIFs generated using the proposed method received 1.02 higher overall ratings compared to the SoA method. To the best of our knowledge, this is the first technique that uses LTC to generate artistic media while providing lightweight and high-performance services even on resource constrained devices.
\end{abstract}

% === KEYWORDS ====================================================================
% =================================================================================

%  thumbnail and artistic animated GIFs

% \fi
\IEEEpeerreviewmaketitle

\section{Introduction}\label{intro}
\IEEEPARstart{O}{ver} the past few years, various types of streaming platforms in the form of video on demand (VoD), 360-degree streaming, and live streaming services have become dramatically popular. Compared to traditional cable broadcast that users can view on television, video streaming is ubiquitous and provides viewers with the flexibility of watching video content on various devices. In most cases, such services have vast videos catalogs present for users to browse and watch anytime. It is often challenging for users to find relevant content due to innumerable data and time constraints. This considerable growth has increased the need for technologies that enable users to browse the vast and ever-growing content collections and quickly retrieve the content of interest. The development of new techniques for generating animated graphic change format (GIF) images and artistic static thumbnails is part of this demand \cite{song2016click, yuan2019sentence, xu2021gif}.

\begin{figure}[t]
\centering
\includegraphics[width=\linewidth,keepaspectratio]{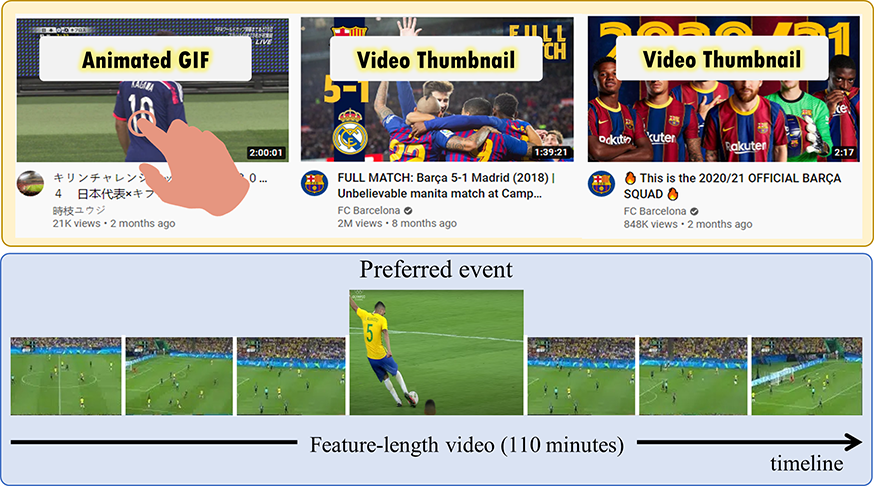}
\caption{\label{fig:intro_gif} Artistic media in the form of static thumbnails and animated GIFs are universally used in the most popular streaming platforms to highlight recommended videos. Animated GIFs are played whenever a user hovers over the static thumbnail (above). Generally, the most preferred events are selected according to the video category to attract users to get more views of the video (below).}
\end{figure}

Almost every streaming platform uses artistic media to provide a quick and decisive glimpse of video content. The artistic static thumbnail provides viewers with a quick video preview. Meanwhile, the animated GIF provides a condensed preview of the video for 3--15 seconds \cite{bakhshi2016fast}. Figure \ref{fig:intro_gif} illustrates artistic media for sports videos: 1) an animated GIF played when the user hovers the mouse on artistic thumbnail (above) and 2) the most preferred frames is selected as artistic thumbnail from the feature-length video. Viewers often decide whether to watch or skip the video based on its static thumbnail and animated GIF. Due to their importance, there is a growing interest in automatically creating compelling and expressive artistic media.

% =====================
% FIG. 01
% =====================

% Clickbait
Click-through rate (CTR) is a prominent metric to boost the popularity of newly published feature-length videos on streaming platforms. However, many streaming platforms (such as YouTube) provide only one thumbnail and a single GIF for a given video, without prioritizing user preferences. Recent studies showed that personalized artistic media (thumbnails and animated GIFs) could play a significant role in video selection and improve the CTR of videos \cite{mujtaba2019client, mujtaba2021GIF}. However, manually creating static thumbnails and GIF thumbnails is time-consuming, and their quality is not guaranteed. Their ubiquitous adoption and prevalence have increased the demand for methods that can automatically generate personalized static artistic media from feature-length videos.

\begin{figure}[t]
\centering
\includegraphics[width=\linewidth,keepaspectratio]{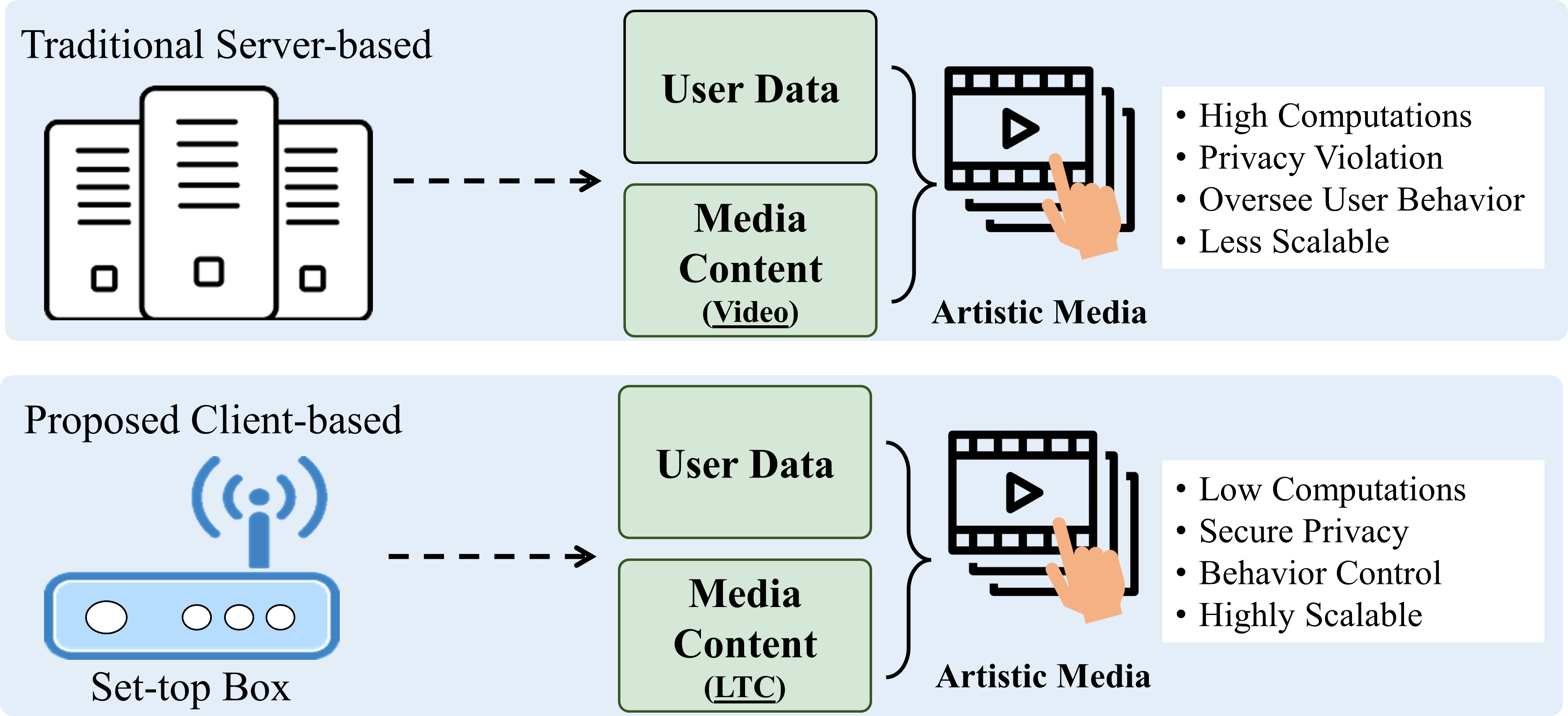}
\caption{\label{fig:2} Traditionally, personalized artistic media (thumbnail and GIF) is generated using server-based techniques. We propose a new lightweight technique in this paper to create personalized artistic media on the client device.}
\end{figure}

Nowadays, some popular video streaming sites are investigating server-side solutions to automatically generate personalized artistic media. There are four key concerns when it comes to server-based solutions: (i) due to finite computing capabilities personalized artistic media may not be simultaneously generated in a timely manner for multiple users, (ii) consumer privacy is prone to invasions in a personalized approach, (iii) user behavior should be overseen with recommendation algorithms, (iv) the fact that the current solution processes the entire video (frames) to generate GIFs increases the overall computational duration and requires significant computational resources. As personalization is one of the key elements for early media content adoption, we focused on the personalization and lightweight processing aspects of artistic media generation. Figure \ref{fig:2} shows a general overview and comparison of the traditional and proposed methods.

% =====================
% FIG. 02
% =====================

With the observation above in mind, we propose an innovative computationally efficient client-driven method that can generate personalized artistic media simultaneously for multiple users. Considering that computational resources are limited, we use lightweight thumbnail containers (LTC) of the corresponding feature-length sports video instead of processing the entire video (frames). Since every sports video has key events (i.e., penalty shots in soccer videos), we utilize LTC to detect events that reduce the overall processing time. Therefore, we aim to reduce the overall computation load and processing time while generating personalized thumbnails and GIFs from feature-length videos. In the proposed method, twenty-three publicly broadcast soccer videos were examined to estimate the model effectiveness\footnote{Here, we focused on long videos of six different sports matches, namely, baseball, basketball, boxing, cricket, football, and tennis. However, the proposed method can also be used for other sporting events}. The main contributions of this research are summarized as follows:

\begin{itemize}
\item We propose a new lightweight client-driven technique to automatically create static artistic media for feature-length sports videos. To the best of our knowledge, this is the first work to address this novel and challenging problem in the literature.
\item To support the study, we have collected twenty-three feature-length videos with approximately $2,818.96$ minutes duration, in six different sports categories, namely, baseball, basketball, boxing, cricket, football, and tennis. 
\item We designed an effective 2D Convolutional Neural Network (CNN) model that can detect personalized events from feature-length videos.
\item Extensive quantitative and qualitative analyses were conducted using feature-length sports videos. The quantitative results indicated that the computational complexity of the proposed method is 3.57 times lower than that of the SoA approach on resource-constrained Nvidia Jetson TX2 device (detailed in Section \ref{sec:level4.3}). Additionally, qualitative evaluations were conducted in collaboration with nine participants (detailed in Section \ref{sec:level4.4}).
\end{itemize}

To the best of our knowledge, this is the first attempt to generate artistic media using LTC in end-user devices for streaming platforms\footnote{The code and trained models are publicly available on GitHub at \url{https://github.com/iamgmujtaba/LTC-GIF}.}.

The rest of this paper is organized as follows: Section II provides an overview of related literature. Section III details the proposed client-driven method. Section IV presents the qualitative and quantitative results, along with the relevant discussions. Finally, the conclusions of this study are presented in Section V.

\section{Related Work}\label{sec:level2}
This paper focuses on artistic media generation methods, event recognition, and video analysis. This section briefly reviews works associated with these topics.

\subsection{Animated GIF Generation Methods}\label{sec:level2.1}
Animated GIFs, first created in 1987, have been widely used in recent years. Specifically, in \cite{bakhshi2016fast}, animated GIFs were reported to be more attractive than other forms of media, including photos and videos, on social media platforms such as Tumblr. They identified some important factors that contribute to fascination users with GIFs, such as animations, storytelling capabilities, and emotional expression. In addition, several studies \cite{chen2017gifgif+, jou2014predicting} have trained models for predicting viewers’ perceptual sentiments toward animated GIFs. Despite the engagement, in \cite{jiang2018perfect}, it was discovered that viewers may have diverse interpretations of animated GIFs used in communication. They predicted facial expressions, histograms, and aesthetic features and compared them to \cite{jou2014predicting} to find the most appropriate video features for expressing useful emotions in GIFs. Another new approach \cite{liu2020sentiment}, sentiment analysis was used to estimate annotated GIF text and visual emotion scores. From an aesthetic perspective, in \cite{song2016click}, frames were picked by measuring various subjective and objective metrics of the video frames (such as visual quality and aesthetics) to generate the GIFs. In a recent study \cite{mujtaba2021GIF}, the authors proposed a client-driven method to mitigate privacy issues while designing a lightweight method for streaming platforms to create GIFs. Instead of adopting full-length video content in the method, the author used an acoustic feature to reduce the overall computational time for resource-contained devices.

% Several researchers have collected and prepared datasets for annotating animated GIFs \cite{gygli2016video2gif}. Researchers have designed the Video2GIF dataset for highlighting videos and extended it to include emotion recognition \cite{gygli2016analyzing}. Another dataset, Image2GIF, has been proposed for video prediction, along with a method to generate a cinemagraph from a single image by predicting future frames \cite{zhou2018image2gif}.

\subsection{Event Recognition Methods}\label{sec:level2.2}
Event recognition is a common problem in detecting and classifying video segments according to the predefined set of actions or activity classes used to understand videos. Most methods adopt temporal segments \cite{yang2019exploring} to prune and classify videos. Recent research has focused on exploiting the context to further improve event recognition. Context represents and utilizes both spatio-temporal information and attention, which helps in learning adaptive confidence scores to utilize surrounding information \cite{heilbron2017scc}. More advanced methods of time integration and motion-aware sequence learning have used other neural networks such as long short-term memory (LSTM) and recurrent neural networks (RNNs) \cite{agethen2019deep, pei2017temporal}. The LSTM convolutional network is designed in combination with attention-based mechanisms to support multiple convolutional kernels and layers. Attention models have also been used to improve the integrated spatio-temporal information. Recent studies have used two model-based attention mechanisms within the analysis of spatio-temporal method \cite{peng2018two}. The first is a spatial level attention model that determines critical areas within a frame, while the second addresses the time level-level attention used to identify frames in a video.

\subsection{Video Understanding Methods}\label{sec:level2.3}
Understanding videos is a prominent field in computer vision research. Event (action) recognition \cite{carreira2017quo} and temporal event localization \cite{farha2019ms} are the two main issues addressed in the literature pertaining to video understanding. Action recognition involves recognizing an action from a cropped video clip, which is accomplished via various methods such as two-stream networks \cite{simonyan2014two}, 3D CNNs \cite{tran2015learning}, and RNNs \cite{donahue2015long}. Another popular action recognition method uses a two-stream structure to extend 3D CNNs \cite{carreira2017quo}. It is obtained by pretraining a 2D CNN model using the ImageNet \cite{deng2009imagenet} dataset and extending the 2D CNN model to a 3D CNN by repeated weighting in a depth-wise manner. These features are local descriptors that are obtained using the bag-of-words method or global descriptors retrieved by CNNs.

Comparing to the proposed method, HECATE \cite{song2016click} is most similar approach as it can generate artistic media. Lightweight client-driven techniques to generate artistic media are still in early stages of development, and more effective methods are needed to bridge the semantic gap between video understanding and personalization. Most modern client devices have limited computational capabilities. Moreover,  inspecting a full-length video to create artistic media is time-consuming and not reasonable for real-time solutions \cite{song2016click}. This paper proposes an effective artistic media generation scheme that considers user preferences and resource-constrained devices. The following section explains the artistic media generation process in detail.

% =====================
% FIG. 04
\begin{figure}[t]
\centering
\includegraphics[width=\linewidth,keepaspectratio]{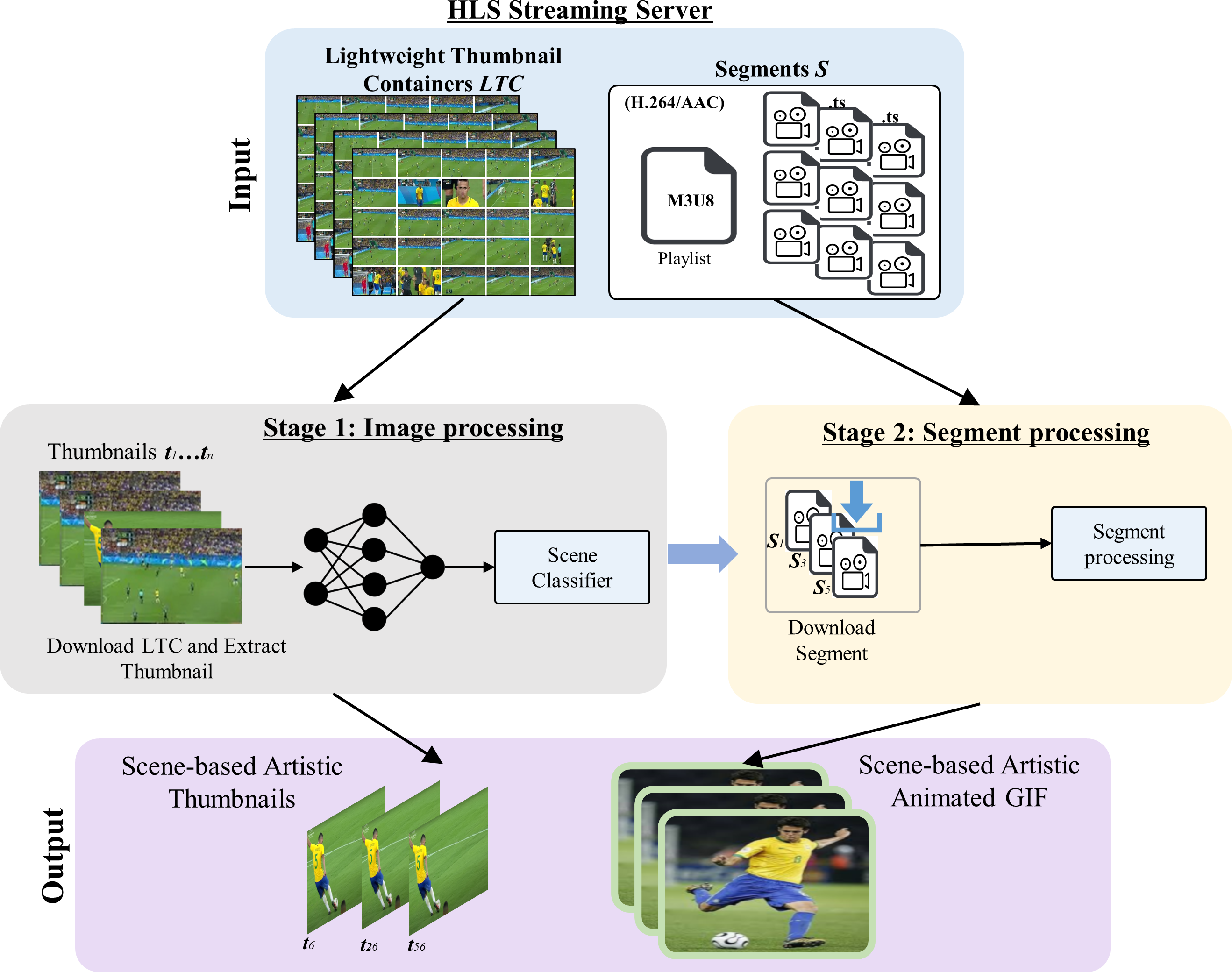}
\caption{\label{fig:propsed_framwework} High-level system architecture of the proposed client-driven LTC artistic media generation method.}
\end{figure}
% =====================

\section{Proposed Method}\label{sec:level3}
\begin{figure}[t]
\centering
\includegraphics[width=\linewidth, height=3.5cm]{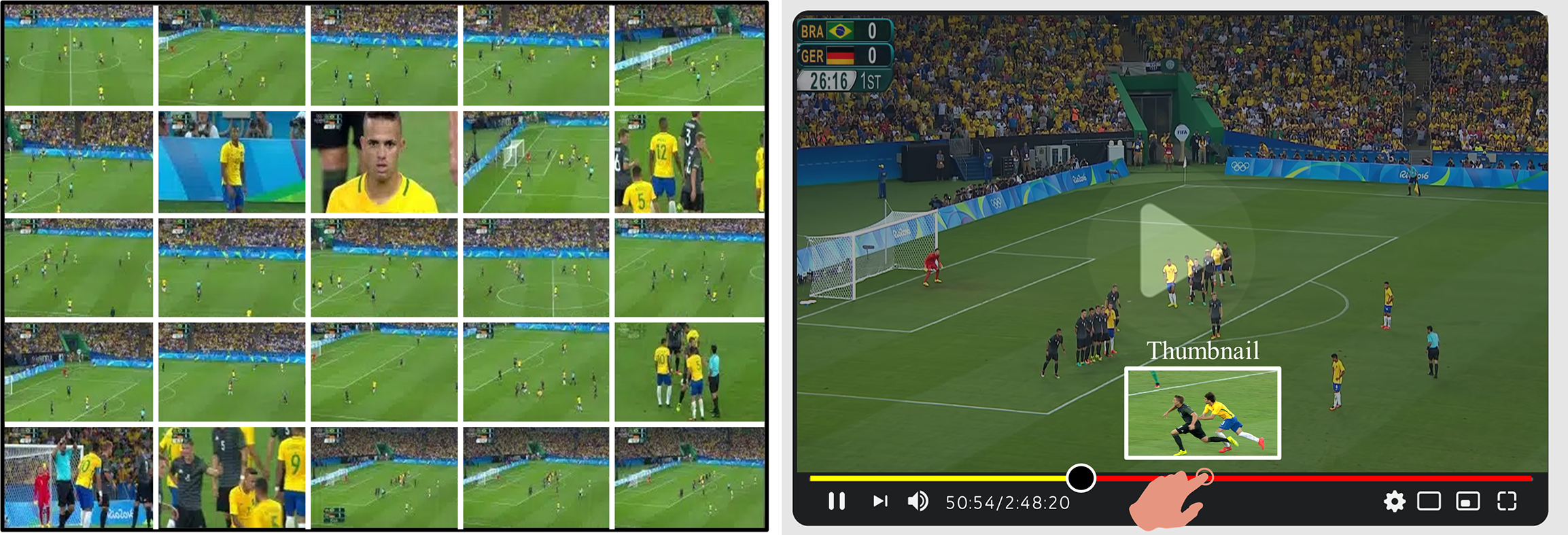}
\caption{\label{fig:3} Example of thumbnail container of selected video (left) and using the thumbnail to instantly preview lengthy videos in web-based players (right).}
\end{figure}

According to a recent study \cite{cisco2020cisco}, the use of streaming platforms has become more popular than ever compared to traditional platforms. CTR is a significantly important metric for streaming platforms, especially for the videos newly broadcast on the platform. Meanwhile, artistic media is vital for streaming platforms as well. There is a stronger correlation between the artistic media and personalization; if the artistic media is relevant to the video, there will be a higher click rate. Consequently, artistic media has become increasingly important in the video selection process. However, currently, they are generated via a one-size-fits-all framework, without user feedback. It is possible that users do not like a particular artistic media because it is not congruent with their interests, which can lead to users skipping the video and reduce its CTR significantly. Owing to the recent popularity of artistic media on streaming platforms, a need for methods that create artistic media based on user preferences with minimal computational requirements has emerged. This paper proposes a new technique to advance the research on generating anticipated artistic media using a client-driven approach. The proposed method uses LTC \footnote{Thumbnail containers are being widely used in streaming platforms for timeline manipulation of videos (refer to Figure \ref{fig:3}) \cite{mujtaba2020client}. This thumbnail container can be obtained from \url{https://www.youtube.com/watch?v=kn5uevla61U}.} instead of the entire video to analyze personalized events. Subsequently, artistic media is created within an adequate processing duration for client-side devices such as Nvidia Jetson TX2, an embedded AI computing device.

Figure \ref{fig:propsed_framwework} depicts the high-level system architecture of the proposed artistic media method. In the streaming server, the size and orientation of the LTC and video segments are identical to those mentioned in a previous work \cite{mujtaba2020client}. There are two phases of generating artistic media. Each phase processes and generates a different artistic media type. In the first phase, LTC is analyzed using the \textit{Thumbnail Container Analyzer} module and artistic thumbnails are obtained. The information in the first phase is used to generate the artistic animated GIF from the given video segment in the second phase of the proposed method. The second phase consists of the \textit{Animated GIF Generation} module. The proposed method and its relevant components are described in the following subsections.

% =====================
% FIG. 03
% =====================

% =====================
% FIG. 05
\begin{figure*}[t]
\centering
\includegraphics[keepaspectratio, width = 17.5cm]{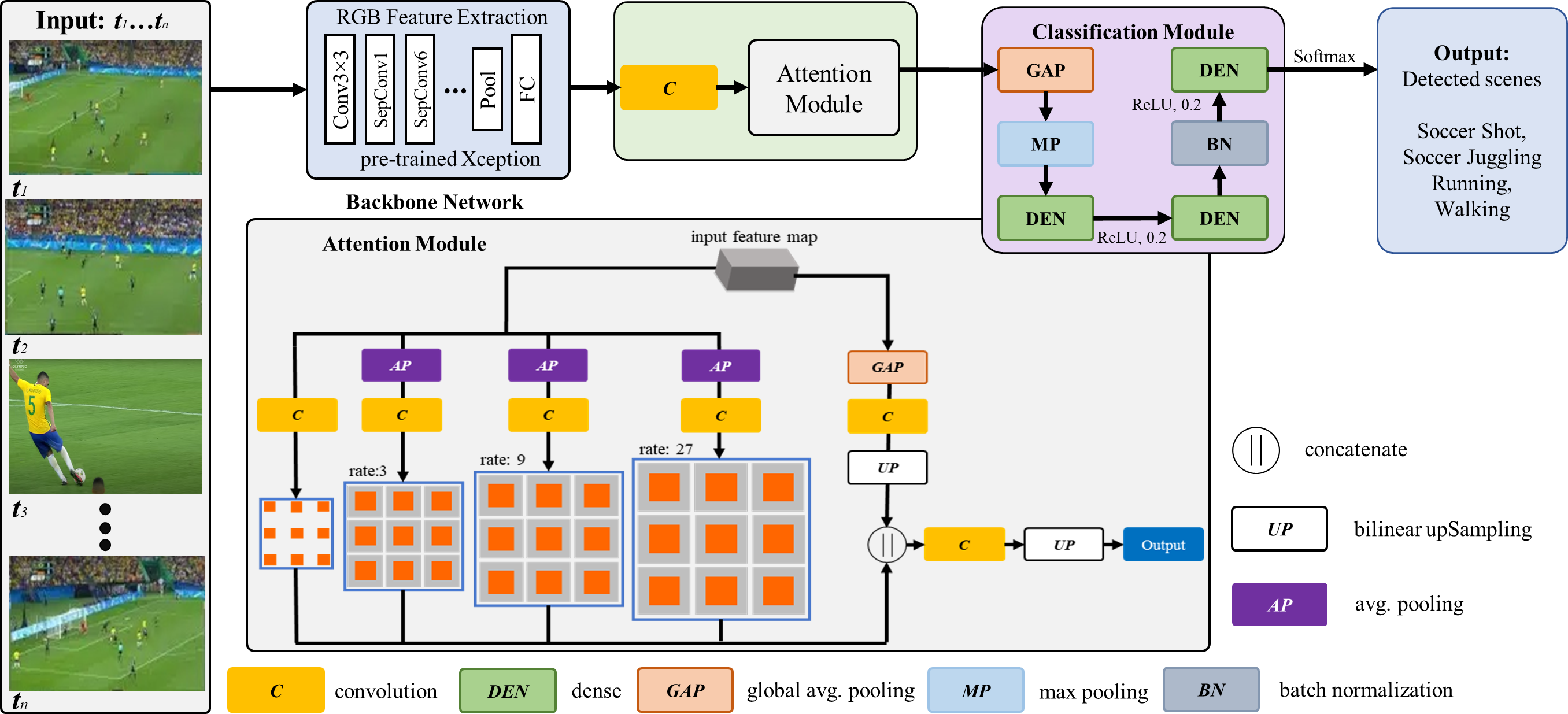}
\caption{\label{fig:deep_lerning} Architecture of the proposed 2D convolutional neural network.}
\end{figure*}
% =====================

% Thumbnail Containers Analyzer
% =======================================================================
\subsubsection{Thumbnail Containers Analyzer Module} \label{sec:level3.2.2}
The LTC analyzer module determines the personalized events from thumbnail containers. A 2D CNN model was designed to examine thumbnails trained on the UCF-101 dataset \cite{soomro2012ucf101}. The dataset was categorized into 101 different action categories from 13,320 videos. The state-of-the-art Xception image annotation model was used to extract frame-level features \cite{chollet2017xception}. The model was pre-trained on ImageNet dataset \cite{deng2009imagenet}. The architecture of the proposed 2D CNN is depicted in Figure \ref{fig:deep_lerning}. Vortex pooling was used as an attention module to enhance the efficiency of the proposed neural network \cite{xie2018vortex}. The module uses multi-branch convolution with dilation rates to aggregate contextual information, making it more effective.

Data augmentation was applied to reduce the overfitting in the proposed approach. The first train/test partition of the UCF-101 dataset was used as recommended in \cite{soomro2012ucf101}. Each video was subsampled up to 40 frames to train the model using the UCF101 dataset. Before being utilized as the network input, all images were pre-processed by cropping their central area and resizing them to 244$\times$244 pixels. Shear transformations were also performed according to an angle of 20°, random rotation of 10°, horizontal and vertical shift of 0.2, and random horizontal inversion of the image. The varied stochastic gradient descent optimizer was used with a learning rate of 0.01, momentum of 0.9, and default weight decay value (SGDW) to train the model \cite{loshchilov2017decoupled}. In the experiment, an early stop mechanism was applied during the training process with a patience of ten. Training data were provided in mini-batches with a size of 32 and a learning rate of 0.001 to minimize costs; 1,000 iterations were performed to train the sequence patterns in the data. The Keras toolbox was used for deep feature extraction, and a GeForce RTX 2080 Ti GPU was used for implementation. Section \ref{sec:level4.2} provides a detailed accuracy analysis of the proposed action recognition model.

% Animated GIFs Generation Module
% =======================================================================
\subsubsection{Animated GIFs Generation Module} \label{sec:level3.2.3}
The animated GIF creation module is designed to examine the segment number from the text-based file generated from detected thumbnails. Later, we have utilized this information to obtain the corresponding segment from the HTTP Live Streaming (HLS) server and create an animated GIF \cite{mujtaba2020client}. The proposed method uses the first 3 seconds of the segment in the animated GIF generation process. FFmpeg is used in the proposed method to create GIFs from segments \cite{ffmpeg}. Here, the duration of all generated GIFs is fixed. However, this approach is extendable to generate a GIF with a specific length. Section \ref{sec:level4.1.2} provides a detailed description of the GIF generation using the proposed method.

% Experimental Results and Discussion
% =======================================================================
\section{Experimental Results and Discussion}\label{sec:level4}
In this section, we present an extensive experimental evaluation of the baseline and proposed approaches. First, the hardware configurations are explained. Next, the entire artistic media process is described from the user's perspective. Later, the experimental scheme is explained with baseline methods. The accuracy of the proposed event recognition model is then given by comparing its performance to those of the prominent action recognition methods on UCF101 dataset. Next, the proposed and baseline methods are compared qualitatively and quantitatively. Finally, the overall results of the proposed and baseline methods are discussed.

% === 5. Experimental Setup =======================================================
% =================================================================================
\subsection{Experimental Setup}\label{sec:level4.1}

\subsubsection{Hardware Configuration}\label{sec:level4.1.1}
The HLS server and HLS client hardware devices were configured locally for the experimental evaluations. For HLS clients, two end-user devices were configured with different hardware configurations: a high computational resource (HCR) end-user device running on the open-source Ubuntu 18.04 LTS operating system, and a low computational resource (LCR) end-user machine utilizing an Nvidia Jetson TX2 device. The proposed and baseline approaches were set up separately on HCR and LCR machines. The HLS server machine was set up with Windows 10 operating system and was used in experiments. The current network structure of our university (SKKU) was utilized to connect all hardware machines locally. Table \ref{tab:hardware_specs} shows the specifications of the hardware devices used in all experiments. The complete artistic media creation process that uses the proposed approach is described in the next subsection from the user's perspective.

% =====================
% Table : 
\begin{table}[ht]
\centering
\caption{\label{tab:hardware_specs} HLS server and HLS clients hardware devices specifications.}
\begin{tabular}{|P{20pt}|P{80pt}|P{75pt}|c|}
\hline
Device & CPU & GPU & RAM \\ 
\Xhline{3\arrayrulewidth}
HLS Server & Intel Core i7-8700K & GeForce GTX 1080 & 32 GB \\ \hline
HCR Client & Quad-core 2.10 GHz Xeon & GeForce RTX 2080 Ti & 62 GB \\ \hline
LCR Client & HMP Dual Denver 2/2MB L2 + Quad ARM A57/2MB L2 & Nvidia Pascal 256 CUDA cores & 8 GB\\ \hline
\end{tabular}
\end{table}
% =====================

% GIF generation process
% =======================================================================
\subsubsection{Proposed Artistic Media Generation Process}
\label{sec:level4.1.2}
This section describes the entire process of artistic media generation from the user's perspective. The process is demonstrated utilizing twenty-three feature-length sports videos obtained from the streaming platform YouTube. The videos are split into six categories based on their content, namely, baseball, basketball, boxing, cricket, football, and tennis. Table \ref{tab:video_title} provides the complete descriptions of the selected videos. View statistics counts were collected in November 2021. All videos used in the experiments had a resolution of $640\times480$ pixels. All selected videos were examined using ten different events selected from the action list provided in the UCF-101 dataset\footnote{It should be noted that the proposed method is not bound by these events; additional events can be included according to the video content.}. The ten selected events were basketball, basketball dunk, boxing punching bag, boxing speed bag, cricket bowling, cricket shot, punch, soccer juggling, soccer penalty, and tennis swing. These events were selected based on the video content.

% SoccerJuggling, SoccerPenalty, Basketball, BasketballDunk, BoxingPunchingBag, BoxingSpeedBag, Punch, CricketBowling, CricketShot, TennisSwing
% 1.	Basketball,
% 2.	BasketballDunk,
% 3.	BoxingPunchingBag,
% 4.	BoxingSpeedBag,
% 5.	CricketBowling,
% 6.	CricketShot,
% 7.	Punch,
% 8.	SoccerJuggling,
% 9.	SoccerPenalty,
% 10.	TennisSwing

% ==========================================
\begin{table*}[t]
\centering
\caption{List of selected videos utilized for analysis in the proposed approach.}
\label{tab:video_title}
\begin{tabular}{|c|c|l|c|c|c|c|c|c|c|}
\hline
S/N & Category                    & \multicolumn{1}{c|}{Title}                & Playtime   & FPS & \# Frames & \# LTC & \# Thumbnails & Views      & YouTube ID   \\ 
\Xhline{3\arrayrulewidth}

1   & \multirow{6}{*}{Football}   & \begin{tabular}[c]{@{}l@{}} Belgium vs. Japan \end{tabular}                           & 1h 52m 14s & 30  & 202,036   & 270   & 6734          & 1,141,707  & ervkVzoFJ5w  \\ \cline{1-1} \cline{3-10} 
2   &                             & \begin{tabular}[c]{@{}l@{}} Brazil vs. Belgium  \end{tabular}                        & 1h 50m 50s & 30  & 199,506   & 267   & 6650          & 935,399    & 5OJfbYQtKtk  \\ \cline{1-1} \cline{3-10} 
3   &                             & \begin{tabular}[c]{@{}l@{}} France vs. Argentina \end{tabular}                       & 1h 50m 26s & 25  & 165,653   & 266   & 6626          & 2,660,920  & J41d0cHAfSM  \\ \cline{1-1} \cline{3-10} 
4   &                             & \begin{tabular}[c]{@{}l@{}} France vs. Croatia  \end{tabular}                        & 1h 54m 1s  & 30  & 205,243   & 274   & 6841          & 1,367,451  & 7Fau-IwbuJc  \\ \cline{1-1} \cline{3-10} 
5   &                             & \begin{tabular}[c]{@{}l@{}} Germany vs. Mexico  \end{tabular}                        & 1h 48m 56s & 30  & 196,106   & 262   & 6536          & 1,111,419  & 3fYpcapas0k  \\ \cline{1-1} \cline{3-10} 
6   &                             & \begin{tabular}[c]{@{}l@{}} Portugal vs. Spain  \end{tabular}                        & 1h 50m 25s & 30  & 198,556   & 266   & 6625          & 1,792,000  & Xhu5Bz1xDf0  \\ \hline

7   & \multirow{4}{*}{Basketball} & \begin{tabular}[c]{@{}l@{}}France vs USA    \end{tabular}                          & 2h 14m 39s & 30  & 242,135   & 324   & 8079          & 1,171,512  & 8YSrNfcKvA0  \\ \cline{1-1} \cline{3-10} 
8   &                             & \begin{tabular}[c]{@{}l@{}}Golden State Warriors   vs. \\ Brooklyn Nets \end{tabular}  & 1h 40m 52s & 30  & 181,574   & 243   & 6052          & 585,904    & KAZ-U8vYqZg  \\ \cline{1-1} \cline{3-10} 
9   &                             & \begin{tabular}[c]{@{}l@{}}Los Angeles Lakers vs.   \\Houston Rockets \end{tabular}    & 1h 54m 19s & 30  & 205,586   & 275   & 6859          & 312,224    & aHVd9vVWVSQ  \\ \cline{1-1} \cline{3-10} 
10  &                             & \begin{tabular}[c]{@{}l@{}}USA vs. Spain - \\Men's   Gold Final   \end{tabular}       & 2h 53m 54s & 25  & 260,886   & 418   & 10434         & 17,722,044 & l9wUr-CK1Y4  \\ \hline
11  & \multirow{4}{*}{Boxing}     & \begin{tabular}[c]{@{}l@{}}Canelo vs. Daniel   Jacobs     \end{tabular}             & 53m 55s    & 30  & 96,968    & 130   & 3235          & 11,834,396 & 1VbXe9ZjzTM  \\ \cline{1-1} \cline{3-10} 
12  &                             & \begin{tabular}[c]{@{}l@{}}Davis vs. Gamboa Full   Fight    \end{tabular}           & 1h 3m 2s   & 30  & 113,368   & 152   & 3782          & 3,135,174  & KZtVQo8lpqY  \\ \cline{1-1} \cline{3-10} 
13  &                             & \begin{tabular}[c]{@{}l@{}}Dirrell vs. Davis Full   Fight    \end{tabular}          & 47m 29s    & 30  & 85,392    & 114   & 2849          & 165,015    & sVtzzpvaEjc  \\ \cline{1-1} \cline{3-10} 
14  &                             & \begin{tabular}[c]{@{}l@{}}Floyd Mayweather Jr.   vs. \\Marcos Maidana \end{tabular} & 56m 50s    & 25  & 85,259    & 137   & 3410          & 13,569,484 & KYvOC7MBuUw  \\ \hline
15  & \multirow{3}{*}{Baseball}   & \begin{tabular}[c]{@{}l@{}}Giants vs.   Dodgers \end{tabular}                     & 2h 11m 42s & 30  & 236,827   & 317   & 7902          & 168,309    & ScmHL8YVM5E  \\ \cline{1-1} \cline{3-10} 
16  &                             & \begin{tabular}[c]{@{}l@{}}Giants vs. Royals \end{tabular}                         & 2h 36m 50s & 30  & 282,024   & 377   & 9410          & 6,448,368  & YJmwofDYOeo  \\ \cline{1-1} \cline{3-10} 
17  &                             & \begin{tabular}[c]{@{}l@{}}Toronto Blue Jays vs.   \\Boston Red Sox \end{tabular}   & 2h 40m 50s & 30  & 289,221   & 387   & 9650          & 19,006     & psL-FvRg9jM  \\ \hline
18  & \multirow{2}{*}{Cricket}    & \begin{tabular}[c]{@{}l@{}}India vs. Pakistan \end{tabular}                         & 1h 25m 2s  & 30  & 153,065   & 205   & 5102          & 36,562,893 & uSGCAJS6qWg  \\ \cline{1-1} \cline{3-10} 
19  &                             & \begin{tabular}[c]{@{}l@{}}Peshawar Zalmi vs.  \\ Islamabad United \end{tabular}      & 2h 17m 15s & 30  & 205,170   & 274   & 6845          & 372,182    & uzErZgKuuSM  \\ \hline
20  & \multirow{4}{*}{Tennis}     & \begin{tabular}[c]{@{}l@{}}Maria Sharapova vs.   \\Caroline Wozniacki \end{tabular}   & 2h 10m 6s  & 30  & 233,962   & 313   & 7806          & 745,690    & 72VhC9biEFk  \\ \cline{1-1} \cline{3-10} 
21  &                             & \begin{tabular}[c]{@{}l@{}}Novak Djokovic vs.   \\Daniil Medvedev  \end{tabular}      & 2h 1m 6s   & 25  & 181,654   & 291   & 7266          & 902,442    & MG-RjlqyaJI  \\ \cline{1-1} \cline{3-10} 
22  &                             & \begin{tabular}[c]{@{}l@{}}Novak Djokovic vs. \\  Roger Federer \end{tabular}         & 4h 58m 38s & 25  & 447,961   & 717   & 17918         & 4,841,514  & TUikJi0Qhhw  \\ \cline{1-1} \cline{3-10} 
23  &                             & \begin{tabular}[c]{@{}l@{}}Roger Federer vs.   Rafael Nadal \end{tabular}          & 3h 5m 37s  & 25  & 278,448   & 446   & 11137         & 4,991,304  & wZnCcqm\_g-E \\ \hline
\end{tabular}
\end{table*}
% ==========================================

To obtain artistic media for a specific video, the user first selects the video from the web interface. The end-user device requests and downloads the LTC for the corresponding video. The downloaded LTC covers the entire duration of the video. The total number of frames, frames per second (FPS), LTC and thumbnails corresponding to LTC in a video are shown in Table \ref{tab:video_title}. A single LTC contains 25 thumbnails. The size of an LTC is considerably smaller than the number of frames in a video; hence, a significantly low bitrate is required during transmission. The proposed method uses the canvas to capture every thumbnail separately from the transmitted thumbnail containers. The event(s) of the video is selected using the web interface. A user can select more than one event during the GIF generation process. The proposed 2D CNN model requires two inputs during the recognition process: the thumbnail and preferred event.

The deep learning model analyzes each extracted thumbnail individually based on the event(s) selected by the user. The proposed method selects a personalized artistic thumbnail from the analyzed LTC. The artistic thumbnails are selected based on a threshold that is set to maintain the quality of the artistic media. A text-based file is generated for all selected personalized artistic thumbnails obtained from the LTC. This text-based file is used to provide personalized artistic thumbnails according to user preferences regarding the video category. The data inside the artistic thumbnail files are ranked in chronological order. To obtain a specific segment for a selected thumbnail, the text-based file is analyzed to download the segment. Then, the end-user machine then requests specific segments from the HLS server with distinct timestamps. The HTTP server sends these segments immediately in response to the client device request. Subsequently, the segments are adopted to create an animated GIF. FFmpeg \cite{ffmpeg} is used in the proposed method to create an artistic GIF from a given segment. Algorithm \ref{code:generate_GIF} depicts the processing steps required to generate a GIF from a video with the proposed method.

% ---------------------------------------
\begin{algorithm} [t]
\DontPrintSemicolon\SetAlgoLined
\noindent\rule{7.5cm}{0.4pt}

\KwData{Input thumbnail containers}
- N: number of thumbnails \textit{T} inside thumbnail containers \textit{LTC}\;
\SetKw{KwInit}{Initialization:}\KwInit
- Personalize events \textit{P}; Segments \textit{S}; threshold = 80\; 
\normalem
\textbf{Main loop}: \While{ i $<$ (N)}
{
Extract \textit{T} from \textit{LTC}\;
\textit{determineEvents}(\textit{T}, \textit{P}, threshold) \;
Identify the \textit{S} number from text-based file \; 
Download \textit{S} \;
Generate animated GIF from \textit{S} \;
}
\SetKwProg{fn}{Function}{}{}
\fn{determineEvents (T, P, threshold)}{
Analyze \textit{T} as per \textit{P}\;
Select artistic \textit{T} according to threshold\;
Prepare text-file of selected \textit{T} \;

% Save thumbnail container $n_{i}$ \;
}\textbf{return} text-based selected \textit{T} list\;
{}

\KwResult{Generated Artistic Media}
\noindent\rule{7.5cm}{0.4pt}
\ULforem
\caption{\label{code:generate_GIF} Process to analyze personalize events from thumbnail containers to generate artistic media.}
\end{algorithm}
% ---------------------------------------

\subsubsection{Baseline Methods}\label{sec:level4.1.3}
This section describes the baseline methods that are compared to the proposed artistic media generation method. As explained in Section \ref{sec:level2}, some of the well-known approaches use the entire video to generate animated GIFs. The baseline approaches are listed as follows:

\begin{itemize}
\item \textbf{HECATE} \cite{song2016click}: It analyzes atheistic features obtained from video frames. The corresponding video is stored locally on the device. During the process, the frames are extracted, temporarily stored, and then analyzed. This method only supports a fixed duration and number of GIFs. Here, ten artistic thumbnail and GIFs were generated for each video.
\item \textbf{AV-GIF} \cite{mujtaba2021GIF}: It analyzes the entire audio and video files to create animated GIFs. This is the baseline approach described in \cite{mujtaba2021GIF}. To create a GIF, the default parameters were used as described by the authors. With this method, only one GIF was generated for each corresponding video using the default parameters.
\item \textbf{CL-GIF} \cite{mujtaba2021GIF}: It uses acoustic features to analyze the audio climax portion and employs segments to generate GIFs. This is the SoA client-driven animated GIF generation method. Default parameters were applied to generate the animated GIFs. Here, similar to \cite{mujtaba2021GIF}, only one GIF was generated using default parameters.
\item \textbf{FB-GIF}: Instead of analyzing the LTC, this method uses video frames of the corresponding video to detect personalized scenes. Initially, frames are extracted from the video; then, the 2D CNN model is used to detect the corresponding events from the extracted frames.
\end{itemize}

% Experimental evaluation of music genre classification model
% =======================================================================
\subsection{Experimental Evaluation Action Recognition}\label{sec:level4.2}
This subsection presents an evaluation of the existing 2D CNN approaches using the UCF-101 dataset. To the best of our knowledge, \cite{mujtaba2020client} is the only method that uses thumbnail containers to recognize events, which performed the best on the UCF-101 dataset when using thumbnail containers. The proposed CNN model performed 2.5\% better in terms of validation accuracy compared to \cite{mujtaba2020client}, with 51.32 million floating-point operations per second. The total number of parameters of the proposed CNN model is 25.6 million. The experimental results of the proposed and baseline approaches on the UCF-101 dataset are listed in Table \ref{tab:methods_comparisons}. All CNN models \cite{chollet2017xception, sandler2018mobilenetv2, howard2019searching, huang2017densely, szegedy2016rethinking} were trained on the UCF-101 dataset with similar configurations without adopting an attention module as described in Section \ref{sec:level3.2.2}. The proposed CNN model was used in all experiments to identify personalized events from thumbnails.

% =====================
\begin{table} [ht]
\centering
\setlength{\tabcolsep}{3pt}
\caption{\label{tab:methods_comparisons} Comparisons between the proposed CNN action recognition model and other approaches.}

\begin{tabular}{ l P{120pt} }
\hline
CNN Methods& Overall validation accuracy (\%) \\ 
\Xhline{3\arrayrulewidth}

MobileNetV2 \cite{sandler2018mobilenetv2} & 59.06\%\\ \hline
MobileNetV3Small \cite{howard2019searching} & 68.75\%\\ \hline
MobileNetV3Large \cite{howard2019searching} & 71.88\%\\ \hline
DenseNet121 \cite{huang2017densely} & 65.31\%\\ \hline
InceptionV3 \cite{szegedy2016rethinking} & 61.25\%\\ \hline
% InceptionResNetV2 \cite{szegedy2016inception} & 74.38\%\\ \hline
Karpathy, Andrej, et al. 2014 \cite{karpathy2014large}& 65.40\%\\ \hline
Shu, Yu, et al. 2018\cite{shu2018odn}& 76.07\%\\ \hline
Mujtaba, et al. 2020 \cite{mujtaba2020client}& 73.75\%\\ \hline
Xception \cite{chollet2017xception}& 68.44\%\\ 
% \hline
\Xhline{3\arrayrulewidth}

\textbf{Proposed} & \textbf{76.25\%} \\ 
% \hline
\Xhline{3\arrayrulewidth}
\end{tabular}
\end{table}
% =====================

% Performance analysis of the proposed method
% =======================================================================
\subsection{Performance Analysis of the Proposed Method}\label{sec:level4.3}
In this section, the performance of the proposed LTC artistic media generation method is evaluated with respect to those of the baseline approaches described in Section \ref{sec:level4.1.3}. This performance evaluation was conducted using twenty-three feature-length sports videos (Table \ref{tab:video_title}). The computation time of the proposed method was calculated considering the (i) download thumbnail containers, (ii) obtaining thumbnails by extracting them from the thumbnail containers, (iii) recognizing personalized event(s) from the thumbnails, (iv) selecting artistic static thumbnails that have high accuracy, (v) estimate segment number and download segments, and (v) creating the artistic animated GIFs from these segments. All thumbnails were selected with an accuracy exceeding 80.0\% of the threshold, which was set to maintain the artistic media quality.

% In this section, the performance of the proposed GIF creation approach is evaluated with respect to those of the baseline approaches described in Section \ref{sec:level4.1.3}. This performance evaluation was conducted using six soccer videos (Table \ref{tab:video_title}). The computation time of the proposed method was calculated considering the (i) download thumbnail containers, (ii) obtaining thumbnails by extracting them from the thumbnail containers, (iii) recognizing personalized event(s) from the thumbnails, (iv) estimate segment number and download segments, and (v) creating the GIFs from these segments. All thumbnails were selected with an accuracy exceeding 80.0\% of the threshold, which was set to maintain the GIF quality.

% \footnote{In all the experiments, model loading time was not added when determining the computation time.}

\begin{figure}[t]
\centering
\includegraphics[width=\linewidth, height=4.5cm]{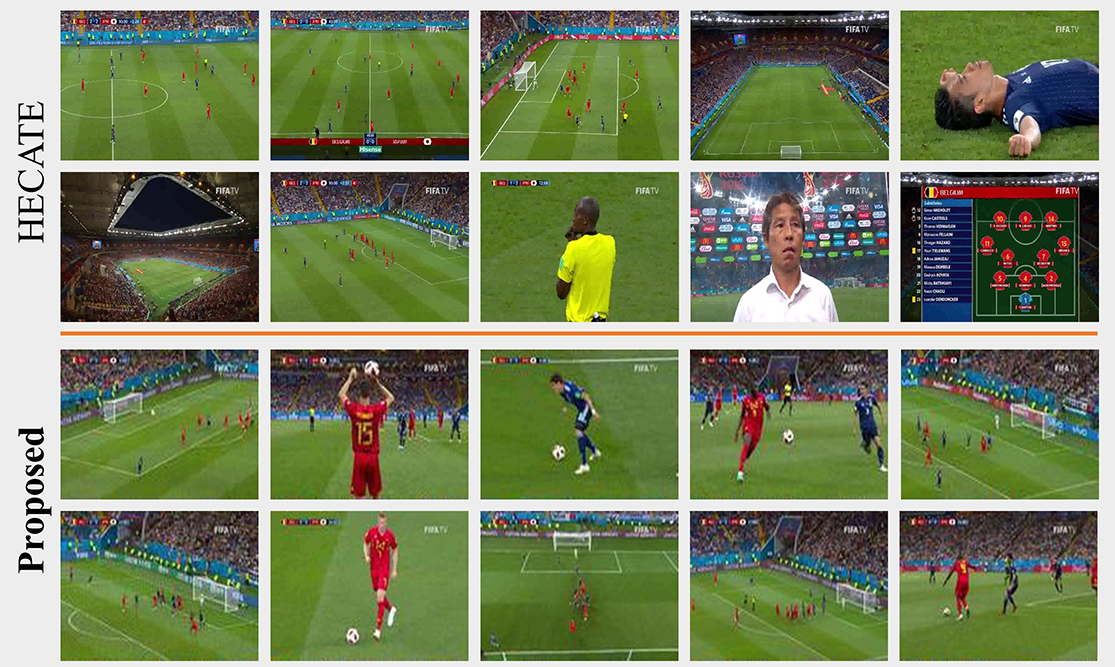}
\caption{\label{fig:thumb_sampel} Artistic thumbnails generated  using proposed and baseline methods.}
\end{figure}

In the first experiment, we evaluated the computation time required to generate artistic static thumbnails using the proposed and baseline approaches. To evaluate the performance of the proposed method, the HECATE \cite{song2016click} baseline method was used with default configuration. In this evaluation, the HCR device was used for experimental evaluation. Table \ref{tab:thumb_hcr} shows number of artistic thumbnails and the computation time required (in minutes) to generate them using proposed and baseline methods. The proposed approach required considerable less computational time then the HECATE \cite{song2016click} baseline method. It is important to note that, all the artistic thumbnails obtained using the proposed method have personalized events. Meanwhile, the artistic thumbnails are generated using HECATE \cite{song2016click} as the one-size-fits-all framework. The artistic thumbnails generated using proposed and baseline methods are depicted in Figure \ref{fig:thumb_sampel}.
% The thumbnails obtained from generated GIFs using the proposed and baseline methods are depicted in Figure \ref{fig:frame_samples}.

% ==================
\begin{table}[t]
\centering
\caption{Computation times required (in minutes) to generate artistic thumbnails using the baseline and proposed methods on the HCR device.}
\label{tab:thumb_hcr}
\begin{tabular}{|c|cc|cc|}
\hline
\multirow{2}{*}{S/N} & \multicolumn{2}{c|}{HECATE \cite{song2016click}}                         & \multicolumn{2}{c|}{Proposed}                              \\ \cline{2-5} 
                     & \multicolumn{1}{c|}{\#Thumbnails} & Total  & \multicolumn{1}{c|}{\#Thumbnails} & Total         \\ \hline
1                    & \multicolumn{1}{c|}{10}                    & 50.19  & \multicolumn{1}{c|}{\textbf{438}}         & \textbf{1.75} \\ \hline
2                    & \multicolumn{1}{c|}{10}                    & 86.59  & \multicolumn{1}{c|}{\textbf{465}}         & \textbf{1.64} \\ \hline
3                    & \multicolumn{1}{c|}{10}                    & 41.34  & \multicolumn{1}{c|}{\textbf{130}}         & \textbf{1.65} \\ \hline
4                    & \multicolumn{1}{c|}{10}                    & 60.17  & \multicolumn{1}{c|}{\textbf{584}}         & \textbf{1.73} \\ \hline
5                    & \multicolumn{1}{c|}{10}                    & 44.78  & \multicolumn{1}{c|}{\textbf{117}}         & \textbf{1.03} \\ \hline
6                    & \multicolumn{1}{c|}{10}                    & 73.16  & \multicolumn{1}{c|}{\textbf{712}}         & \textbf{1.67} \\ \hline
7                    & \multicolumn{1}{c|}{10}                    & 130.16 & \multicolumn{1}{c|}{\textbf{984}}         & \textbf{2.01} \\ \hline
8                    & \multicolumn{1}{c|}{10}                    & 67.40  & \multicolumn{1}{c|}{\textbf{961}}         & \textbf{1.51} \\ \hline
9                    & \multicolumn{1}{c|}{10}                    & 66.35  & \multicolumn{1}{c|}{\textbf{1040}}         & \textbf{1.77} \\ \hline
10                   & \multicolumn{1}{c|}{10}                    & 158.84 & \multicolumn{1}{c|}{\textbf{1184}}         & \textbf{2.64} \\ \hline
11                   & \multicolumn{1}{c|}{10}                    & 14.70  & \multicolumn{1}{c|}{\textbf{928}}         & \textbf{0.82} \\ \hline
12                   & \multicolumn{1}{c|}{10}                    & 19.63  & \multicolumn{1}{c|}{\textbf{845}}         & \textbf{0.96} \\ \hline
13                   & \multicolumn{1}{c|}{10}                    & 13.33  & \multicolumn{1}{c|}{\textbf{897}}         & \textbf{0.72} \\ \hline
14                   & \multicolumn{1}{c|}{10}                    & 14.05  & \multicolumn{1}{c|}{\textbf{1375}}         & \textbf{0.86} \\ \hline
15                   & \multicolumn{1}{c|}{10}                    & 87.20  & \multicolumn{1}{c|}{\textbf{1225}}         & \textbf{1.99} \\ \hline
16                   & \multicolumn{1}{c|}{10}                    & 81.25  & \multicolumn{1}{c|}{\textbf{1020}}         & \textbf{2.36} \\ \hline
17                   & \multicolumn{1}{c|}{10}                    & 78.41  & \multicolumn{1}{c|}{\textbf{1160}}         & \textbf{2.40} \\ \hline
18                   & \multicolumn{1}{c|}{10}                    & 25.29  & \multicolumn{1}{c|}{\textbf{18}}           & \textbf{1.28} \\ \hline
19                   & \multicolumn{1}{c|}{10}                    & 48.70  & \multicolumn{1}{c|}{\textbf{14}}           & \textbf{1.60} \\ \hline
20                   & \multicolumn{1}{c|}{10}                    & 65.19  & \multicolumn{1}{c|}{\textbf{158}}          & \textbf{1.95} \\ \hline
21                   & \multicolumn{1}{c|}{10}                    & 34.41  & \multicolumn{1}{c|}{\textbf{124}}          & \textbf{1.84} \\ \hline
22                   & \multicolumn{1}{c|}{10}                    & 178.28 & \multicolumn{1}{c|}{\textbf{45}}           & \textbf{4.42} \\ \hline
23                   & \multicolumn{1}{c|}{10}                    & 74.86  & \multicolumn{1}{c|}{\textbf{22}}           & \textbf{2.78} \\ \hline
\end{tabular}
\end{table}
% ==================

% =====================
% FIG. 06
% =====================

In the second experiment, we compared the computational time required to generate artistic animated GIFs using the proposed and baseline approaches. The HCR device was adopted to all approaches, and the detailed device specification for each approach are highlighted in Table \ref{tab:hardware_specs}. The computational times required (in minutes) to generate GIFs using the proposed and baseline approaches is depicted in Table \ref{tab:baseline_hcr}. Table \ref{tab:thumb_compare} shows the number of events and segments detected using baseline LTC approach with the proposed method. Table \ref{tab:proposed_hcr} depicts the computation time required (in seconds) for every step when creating artistic media on the HCR device. The HECATE \cite{song2016click} method analyzes every frame in the video and determines aesthetic features that can be used for generating GIFs. The number of thumbnails is significantly lower than the number of frames in the example video, shown in Table \ref{tab:video_title}. As indicated in a recent paper, AV-GIF \cite{mujtaba2021GIF} uses entire video and audio clips to generate animated GIFs. Meanwhile, CL-GIF \cite{mujtaba2021GIF} uses segments and audio climax portions to generate animated GIFs. The proposed method uses considerably small images (thumbnails) to analyze personalized events, which results in a significantly lower computation time for generating animated GIFs.

% ==================
\begin{table}[t]
\centering
\caption{Computation times required (in minutes) to generate artistic animated GIFs using the baseline and proposed methods on the HCR device.}
\label{tab:baseline_hcr}
\begin{tabular}{| c | c | P{30pt} | P{30pt} |c|c|}
\hline
S/N & HECATE \cite{song2016click} & AV-GIF \cite{mujtaba2021GIF} & CL-GIF \cite{mujtaba2021GIF} & FB-GIF & Proposed \\ 
\Xhline{3\arrayrulewidth}
1   & 51.52  & 21.60         & 8.16          & 70.67  & \textbf{2.20} \\ \hline
2   & 89.79  & 21.36         & 8.56          & 65.31  & \textbf{2.02} \\ \hline
3   & 45.69  & 21.09         & 0.77          & 54.81  & \textbf{2.09} \\ \hline
4   & 103.63 & 20.20         & 8.26          & 117.72 & \textbf{2.13} \\ \hline
5   & 45.29  & 22.04         & 8.29          & 63.74  & \textbf{1.49} \\ \hline
6   & 76.34  & 42.88         & 7.66          & 65.60  & \textbf{2.03} \\ \hline
7   & 199.44 & 26.36         & 8.22          & 137.77 & \textbf{2.38} \\ \hline
8   & 97.36  & 16.24         & 7.41          & 127.98 & \textbf{1.90} \\ \hline
9   & 97.86  & 19.14         & 7.86          & 177.38 & \textbf{2.29} \\ \hline
10  & 245.67 & 47.64         & 12.55         & 84.30  & \textbf{3.09} \\ \hline
11  & 16.24  & 9.58          & 3.52          & 42.63  & \textbf{1.22} \\ \hline
12  & 33.12  & 10.86         & 4.87          & 64.33  & \textbf{1.37} \\ \hline
13  & 20.92  & 8.07          & 3.04          & 43.35  & \textbf{1.17} \\ \hline
14  & 14.13  & 10.92         & 3.66          & 29.21  & \textbf{1.36} \\ \hline
15  & 93.92  & 29.68         & 9.38          & 155.61 & \textbf{2.47} \\ \hline
16  & 132.03 & 104.24        & 15.52         & 98.34  & \textbf{2.83} \\ \hline
17  & 88.27  & 30.01         & 13.83         & 94.66  & \textbf{2.85} \\ \hline
18  & 35.08  & 17.38         & 6.68          & 48.44  & \textbf{1.71} \\ \hline
19  & 49.70  & 23.92         & 9.93          & 69.22  & \textbf{2.03} \\ \hline
20  & 79.53  & 31.44         & 10.48         & 90.68  & \textbf{2.34} \\ \hline
21  & 35.18  & 41.99         & 10.98         & 58.26  & \textbf{2.20} \\ \hline
22  & 128.32 & 31.37         & 20.79         & 152.01 & \textbf{4.86} \\ \hline
23  & 79.24  & 41.05         & 13.87         & 181.49 & \textbf{3.18} \\ \hline
\end{tabular}
\end{table}
% ==================

% ==================
\begin{table}[t]
\centering
\caption{The number of detected events from LTC and segments using the proposed method compared to the previous methods.}
\label{tab:thumb_compare}
\begin{tabular}{|c|cP{40pt}|cP{40pt}|}
\hline
\multirow{2}{*}{S/N} & \multicolumn{2}{c|}{Mujtaba, et al. 2020 \cite{mujtaba2020client}}        & \multicolumn{2}{c|}{Proposed Method}               \\ \cline{2-5} 
                     & \multicolumn{1}{P{45pt}|}{Events} & Segments & \multicolumn{1}{P{45pt}|}{Events} &  Segments \\ 
\Xhline{3\arrayrulewidth}
1                    & \multicolumn{1}{c|}{403}            & 203          & \multicolumn{1}{c|}{\textbf{1849}}           & \textbf{417}          \\ \hline
2                    & \multicolumn{1}{c|}{465}            & 211          & \multicolumn{1}{c|}{\textbf{2819}}           & \textbf{491}          \\ \hline
3                    & \multicolumn{1}{c|}{130}            & 82           & \multicolumn{1}{c|}{\textbf{1540}}           & \textbf{389}          \\ \hline
4                    & \multicolumn{1}{c|}{584}            & 223          & \multicolumn{1}{c|}{\textbf{3084}}           & \textbf{499}          \\ \hline
5                    & \multicolumn{1}{c|}{117}            & 71           & \multicolumn{1}{c|}{\textbf{2238}}           & \textbf{447}          \\ \hline
6                    & \multicolumn{1}{c|}{1712}           & 412          & \multicolumn{1}{c|}{\textbf{3926}}           & \textbf{520}          \\ \hline
7                    & \multicolumn{1}{c|}{1082}           & 330          & \multicolumn{1}{c|}{\textbf{2930}}           & \textbf{497}          \\ \hline
8                    & \multicolumn{1}{c|}{2425}           & 417          & \multicolumn{1}{c|}{\textbf{2461}}           & \textbf{421}          \\ \hline
9                    & \multicolumn{1}{c|}{1140}           & 351          & \multicolumn{1}{c|}{\textbf{3912}}           & \textbf{541}          \\ \hline
10                   & \multicolumn{1}{c|}{1184}           & 344          & \multicolumn{1}{c|}{\textbf{3376}}           & \textbf{540}          \\ \hline
11                   & \multicolumn{1}{c|}{1528}           & 242          & \multicolumn{1}{c|}{\textbf{1719}}           & \textbf{270}          \\ \hline
12                   & \multicolumn{1}{c|}{1489}           & 283          & \multicolumn{1}{c|}{\textbf{1341}}           & \textbf{261}          \\ \hline
13                   & \multicolumn{1}{c|}{1149}           & 218          & \multicolumn{1}{c|}{\textbf{1477}}           & \textbf{241}          \\ \hline
14                   & \multicolumn{1}{c|}{1875}           & 274          & \multicolumn{1}{c|}{\textbf{2295}}           & \textbf{301}          \\ \hline
15                   & \multicolumn{1}{c|}{2123}           & 468          & \multicolumn{1}{c|}{\textbf{2599}}           & \textbf{557}          \\ \hline
16                   & \multicolumn{1}{c|}{1619}           & 425          & \multicolumn{1}{c|}{\textbf{2044}}           & \textbf{512}          \\ \hline
17                   & \multicolumn{1}{c|}{1959}           & 535          & \multicolumn{1}{c|}{\textbf{3328}}           & \textbf{692}          \\ \hline
18                   & \multicolumn{1}{c|}{8}              & 4            & \multicolumn{1}{c|}{\textbf{25}}             & \textbf{12}           \\ \hline
19                   & \multicolumn{1}{c|}{10}             & 7            & \multicolumn{1}{c|}{\textbf{22}}             & \textbf{17}           \\ \hline
20                   & \multicolumn{1}{c|}{218}            & 90           & \multicolumn{1}{c|}{\textbf{364}}            & \textbf{134}          \\ \hline
21                   & \multicolumn{1}{c|}{146}            & 62           & \multicolumn{1}{c|}{\textbf{124}}            & \textbf{66}           \\ \hline
22                   & \multicolumn{1}{c|}{56}             & 34           & \multicolumn{1}{c|}{\textbf{82}}             & \textbf{53}           \\ \hline
23                   & \multicolumn{1}{c|}{25}             & 20           & \multicolumn{1}{c|}{\textbf{87}}             & \textbf{51}           \\ \hline
\end{tabular}
\end{table}
% ==================

Since this study focuses on generating artistic media using resource-constrained end-user devices, subsequent experiments are conducted implementing the proposed and baseline methods on the LCR device, namely, Nvidia Jetson TX2. Table \ref{tab:baseline_lcr} shows the computation times required (in minutes) of first six feature-length sports videos to create artistic GIFs when implementing the baseline and proposed methods on the LCR device. In this experiment, we considered HECATE \cite{song2016click} and AV-GIF \cite{mujtaba2021GIF} approaches to generate animated GIFs. However, these approaches cannot be used in practice because they require significant computational resources owing to requiring lengthy videos. Only the CL-GIF \cite{mujtaba2021GIF} method can be used on the LCR device to generate a GIF. The overall processing time of the proposed method is significantly shorter than that of CL-GIF \cite{mujtaba2021GIF}.

% ==================
\begin{table}[ht]
\centering
\caption{Computation times required (in minutes) to generate artistic GIFs using the baseline and proposed methods on the LCR device.}
\label{tab:baseline_lcr}
\begin{tabular}{| P{15pt} | P{92pt} | P{93pt} |}
\hline
S/N & CL-GIF \cite{mujtaba2021GIF} & Proposed \\ 
\Xhline{3\arrayrulewidth}
1 & 38.71& \textbf{10.08} \\ \hline
2 & 36.17& \textbf{9.85}\\ \hline
3 & 35.40& \textbf{9.32}\\ \hline
4 & 40.06& \textbf{10.45} \\ \hline
5 & 37.96& \textbf{13.96} \\ \hline
6 & 35.60& \textbf{8.92}\\ \hline
\end{tabular}
\end{table}
% ==================

From the communication and storage perspectives, the proposed approach is more effective than the baseline methods. The HECATE approach requires a locally stored video file to begin processing \cite{song2016click}. Similarly, the corresponding full-length audio file and video segment must be downloaded when using the CL-GIF method to generate a GIF \cite{mujtaba2021GIF}. However, the proposed method requires only a lightweight thumbnail container downloaded for the same process. For example, the video and audio sizes of the Germany vs. Mexico match were 551 MB and 149 MB, respectively. However, the thumbnail container size was 22.2 MB for the same video. Thus, the proposed method significantly reduced the download time and storage requirements compared to the baseline methods.

% =====================
% FIG. 07
\begin{figure*}[t]
\centering
\includegraphics[keepaspectratio, width=\linewidth]{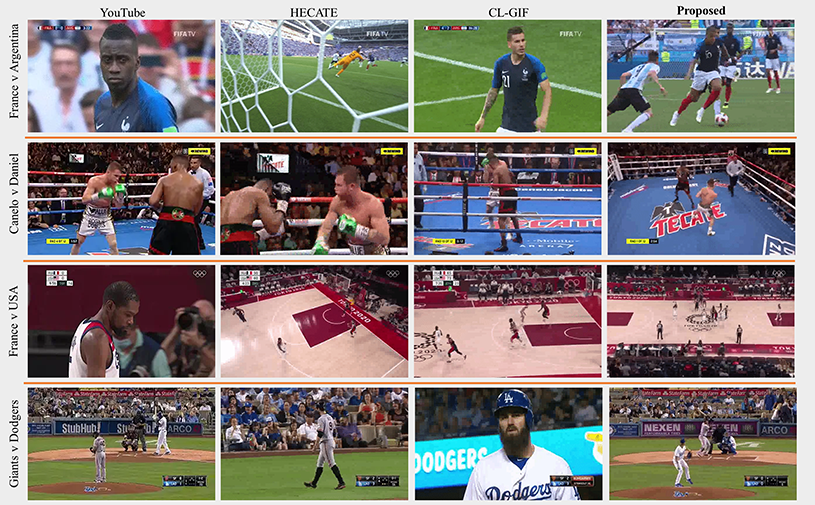}
\caption{\label{fig:frame_samples} Sample frames taken from the GIFs generated using the proposed and baseline methods.}
\end{figure*}
% =====================

The total computation time for the twenty-three feature-length videos was $2,818.96$ minutes. 
To create artistic thumbnails for the corresponding videos using HCR end-user device, HECATE~\cite{song2016click} required $1514.27$ minutes and the proposed method required $41.35$ minutes. Therefore, the analysis of these twenty-three videos indicates that, on average, the proposed method is $36.62$ times faster than the HECATE \cite{song2016click} when generating the personalized artistic thumbnails. 
%To create artistic thumbnails for the corresponding videos using HCR end-user device, \cite{song2016click} needed $1514.27$ minutes and the proposed method requires $41.35$ minutes. Therefore, an analysis of these twenty-three videos indicates that, on average, the proposed method is $36.62$ times faster than the HECATE \cite{song2016click} while generating personalized artistic thumbnails. 
To create the corresponding GIFs when using the HCR end-user device, HECATE \cite{song2016click}, AV-GIF \cite{mujtaba2021GIF}, CL-GIF \cite{mujtaba2021GIF}, FB-GIF, and proposed method required $1858.25$, $649.07$, $204.31$, $2093.52$, and $51.20$ minutes, respectively. Moreover, for the first six videos, CL-GIF \cite{mujtaba2021GIF} and proposed method needed $223.92$ and $62.59$ minutes, respectively, on the LRC device (Table~\ref{tab:baseline_lcr}). Therefore, the analysis of these twenty-three videos indicates that, on average, the proposed method is $36.29$, $40.88$, $12.67$, and $3.99$ times faster than the HECATE \cite{song2016click}, FB-GIF, AV-GIF \cite{mujtaba2021GIF}, and CL-GIF \cite{mujtaba2021GIF} methods when using the HCR device, respectively. Similarly, when using the LCR device, the proposed method is $3.57$ times faster than the CL-GIF \cite{mujtaba2021GIF} method. The proposed approach also generates more GIFs than baseline methods. For example, when generating one GIF with AV-GIF and CL-GIF methods, $10$ GIFs can be generated using HECATE \cite{song2016click}, whereas the proposed method can generate $25$ GIFs.
%To create the corresponding GIFs when using the HCR end-user device, HECATE \cite{song2016click} needed $1858.25$ minutes, AV-GIF \cite{mujtaba2021GIF} required $649.07$ minutes, CL-GIF \cite{mujtaba2021GIF} required $204.31$ minutes, FB-GIF required $2093.52$ minutes, and the proposed method requires $51.20$ minutes. Moreover, for the first six videos CL-GIF \cite{mujtaba2021GIF} needs $223.92$ minutes, and the proposed method needs $62.59$ minutes on the LRC device. Therefore, an analysis of these twenty-three videos indicates that, on average, the proposed method is $36.29$ times faster than the HECATE \cite{song2016click} method, $40.88$ time times faster than the FB-GIF, $12.67$ times faster than the AV-GIF \cite{mujtaba2021GIF} method, and $3.99$ times faster than the CL-GIF \cite{mujtaba2021GIF} technique when using the HCR device. Similarly, for the LCR device, the proposed method is $3.57$ times faster than the CL-GIF \cite{mujtaba2021GIF} method. The proposed approach also generate more GIFs than the baseline method, like, $10$ GIFs generated using HECATE \cite{song2016click}, one GIF for each AV-GIF \cite{mujtaba2021GIF} and CL-GIF \cite{mujtaba2021GIF} method, and $25$ GIFs are generated using the proposed method. 
In summary, these outcomes prove that the proposed approach is more computationally effective than the baseline methods when using both HCR and LCR devices.

% ==================
\begin{table}[!b]
\centering
\caption{Average ratings (1$\sim$10) assigned by participants for the proposed and baseline methods.}
\label{tab:quantitative_eva}
\begin{tabular}{|P{15pt} |c|c|c| P{35pt} |}
\hline
S/N & YouTube & HECATE \cite{song2016click} & CL-GIF \cite{mujtaba2021GIF} & Proposed\\ 
\Xhline{3\arrayrulewidth}
1 & 4.67& 6.78 & 5.67 & \textbf{8.11} \\ \hline
2 & 4.67& 6.22 & 7.00 & \textbf{8.56} \\ \hline
3 & 4.78& 7.56 & 5.33 & \textbf{8.44} \\ \hline
4 & 5.56& 5.44 & 5.22 & \textbf{5.78} \\ \hline
5 & 4.22& 6.33 & 5.00 & \textbf{7.44} \\ \hline
6 & 6.11& 6.44 & 5.67 & \textbf{6.56} \\ \hline
\end{tabular}
\end{table}
% ==================

% Qualitative Evaluation
% =======================================================================
\subsection{Qualitative Evaluation}
\label{sec:level4.4}
This section evaluates the quality of GIFs created using the proposed approach compared to those obtained from YouTube or created utilizing baseline approaches. The evaluation was conducted using a survey with nine participants. A group of students was selected based on their interest in sports. The survey was based on the first six videos (Table \ref{tab:video_title}). The quality of the created GIFs was assessed with respect to exact rating scales. The participants were asked to grade the GIFs based on perceived joy. An anonymous questionnaire was designed for the created GIFs to prevent users from determining the method used to create a given GIF. The participants were requested to view all GIFs and rank them on a scale of 1 to 10 (1 being the lowest and 10 being the highest ranking). Table \ref{tab:quantitative_eva} lists the rankings of the three methods given by the participants.
%The ratings assigned by the participants for all three methods are depicted in Table \ref{tab:quantitative_eva}. 
Regarding the six videos, the average ratings for YouTube, HECATE \cite{song2016click}, CL-GIF \cite{mujtaba2021GIF}, and the proposed method were $5.0$, $6.46$, $5.65$, and $7.48$, respectively. The sample frames obtained from the generated GIFs using the proposed and baseline methods are presented in Figure \ref{fig:frame_samples}.

% ==================
\begin{table*}[t]
\centering
\caption{Computation time required (in seconds) at each step when implementing the proposed method using the HCR device. }
\label{tab:proposed_hcr}
\begin{tabular}{|c|c|c|c|c|c|c|c|}
\hline
S/N & Download LTC & Extract Thumbnails & Events & Thumbnail Selection  & Download Segments & Generate GIFs & Total (sec)\\ 
% \hline
\Xhline{3\arrayrulewidth}

1   & 5.05        & 7.42      & 92.54   & 0.1        & 4.14              & 22.62        & 131.87 \\ \hline
2   & 5.34        & 7.43      & 85.58   & 0.1        & 3.71              & 19.11        & 121.27 \\ \hline
3   & 5.59        & 7.58      & 85.87   & 0.1        & 4.45              & 21.82        & 125.31 \\ \hline
4   & 5.66        & 7.82      & 90.05   & 0.1        & 3.77              & 20.45        & 127.85 \\ \hline
5   & 5.96        & 7.51      & 48.45   & 0.1        & 4.37              & 23.29        & 89.68  \\ \hline
6   & 5.15        & 7.42      & 87.42   & 0.1        & 3.05              & 18.83        & 121.97 \\ \hline
7   & 7.16        & 9.12      & 104.19  & 0.1        & 3.95              & 18.26        & 142.78 \\ \hline
8   & 5.73        & 7.35      & 77.78   & 0.1        & 4.13              & 18.83        & 113.92 \\ \hline
9   & 7.36        & 8.39      & 90.68   & 0.1        & 4.09              & 26.67        & 137.29 \\ \hline
10  & 10.75       & 12.29     & 135.26  & 0.1        & 4.47              & 22.6         & 185.47 \\ \hline
11  & 3.31        & 3.87      & 42.17   & 0.1        & 3.94              & 19.99        & 73.38  \\ \hline
12  & 3.79        & 4.42      & 49.12   & 0.1        & 3.71              & 21.24        & 82.38  \\ \hline
13  & 2.7         & 3.36      & 37.09   & 0.1        & 3.7               & 23.12        & 70.07  \\ \hline
14  & 3.84        & 4.32      & 43.37   & 0.1        & 4.62              & 25.69        & 81.94  \\ \hline
15  & 7.1         & 9.02      & 102.99  & 0.1        & 4.6               & 24.43        & 148.24 \\ \hline
16  & 8.45        & 10.79     & 122.24  & 0.1        & 3.59              & 24.82        & 169.99 \\ \hline
17  & 8           & 10.85     & 124.88  & 0.1        & 2.85              & 24.27        & 170.95 \\ \hline
18  & 3.96        & 5.59      & 67.23   & 0.1        & 2.66              & 23.45        & 102.99 \\ \hline
19  & 5.08        & 7.42      & 83.39   & 0.1        & 2.64              & 23           & 121.63 \\ \hline
20  & 6.81        & 8.9       & 100.99  & 0.1        & 3.94              & 19.59        & 140.33 \\ \hline
21  & 5.82        & 7.84      & 96.69   & 0.1        & 3.33              & 18.13        & 131.91 \\ \hline
22  & 15.7        & 20.74     & 228.8   & 0.1        & 2.86              & 23.62        & 291.82 \\ \hline
23  & 9.61        & 13.16     & 143.79  & 0.1        & 3.74              & 20.61        & 191.01 \\ \hline
\end{tabular}
\end{table*}

% ==================

% Discussion
% =======================================================================
\subsection{Discussion}
\label{sec:level4.5}
The overall effectiveness of the proposed method was evaluated through comparisons to the baseline methods. The proposed method achieved significantly higher performance and shorter computational time on both HCR and LCR devices because it uses thumbnail containers and video segments to generate artistic media instead of processing the entire video, which results in improved computational efficiency. The superiority of our method was underlined experimentally as well, whose results were compared to those of the baseline methods. The proposed method was shown to be $36.62$ times faster than the HECATE \cite{song2016click} while generating artistic thumbnails when using the HCR device. Meanwhile, the proposed method was shown to be $36.29$, $40.88$, $12.67$, and, $3.99$ times faster than the HECATE \cite{song2016click}, FB-GIF, AV-GIF \cite{mujtaba2021GIF}, the CL-GIF \cite{mujtaba2021GIF} methods during artistic animated GIF generation, respectively, when using the HCR device. Similarly, when using the LCR device, the proposed method is 3.57 times faster while analyzing six video than the CL-GIF \cite{mujtaba2021GIF} method. The proposed method has reduced the overall computational power and time required to produce GIFs on client devices.

In the qualitative experiment involving participants, detailed in Section \ref{sec:level4.4}, the proposed approach obtained a higher average rating than the those of other methods. This is mainly because the GIFs are generated based on user interests with the proposed approach. In addition, the proposed method can generate more than one GIF, which can then be used randomly to obtain a greater CTR for the corresponding video. In practical applications, the proposed method can significantly increase the CRT of soccer and other full-length newly broadcast sports videos.

The proposed system can be used a wide range of client devices with different computational resource capabilities. Thanks to its simplicity and scalability in implementing multiple device configurations \cite{li2020energy}, it can be easily adapted to other animated image formats, such as WebP, recommendation methods \cite{mu2020auxiliary, zhang2020social}, and other streaming protocols. In addition, by reducing the computational load of the servers, the proposed approach can act as a privacy protection solution by utilizing effective encryption methods \cite{mujtaba2019, ryu2011home, ryu2008towards} in three-screen TV solutions \cite{kim2019360, jeong2019towards}. Various client-based GIF generation real-time application scenarios for smartphones or set-top boxes can be considered. For an example, if the battery is fully charged, an iPhone utilizes its computing resources to analyze the photos/videos from specific dates and generates the so-called ``memories'' video summary. Animated GIFs can be generated similarly using end-user devices. Client-based GIF generation technology is in early-stages of development and new methods considering different scenarios need to be researched.

%%%%%%%%%%%%%%%%%%%%%%%%%%%%%%%%%%%%%%%%%%
\section{Conclusions}
This paper proposes a new lightweight method for generating artistic media the computational resources of end-user devices. The proposed method analyzes thumbnails to recognize personalized events and uses the corresponding video segments to generate artistic thumbnail and animated GIFs. This improves the computational efficiency and reduces the demand for communication and storage resources in resource-constrained devices. Extensive experimental results based on a set of twenty-three videos show that the proposed approach is 3.99 and 3.57 times faster than the SoA method, respectively when using HCR and LCR devices. The qualitative evaluation indicated that the proposed method outperformed the existing methods and received higher overall ratings. In the future, the proposed method could be implemented for other sports categories by considering various events using resource-constrained devices.


\normalem
\bibliographystyle{IEEEtran}
\bibliography{IEEEabrv,manuscript_r1}

\end{document}